%% file: main.tex
\pdfoutput=1

\documentclass[11pt]{article}

\usepackage[]{main}

\usepackage{times}
\usepackage{latexsym}
\usepackage{multirow}

\usepackage{tablefootnote}

\usepackage{amsmath}
\usepackage{booktabs}
\usepackage{graphicx}
\usepackage{cleveref}

\usepackage{xspace}
\usepackage{color}

\usepackage{caption}
\usepackage{subcaption}
\usepackage{colortbl}

\usepackage{array}

\usepackage[T1]{fontenc}

\usepackage[utf8]{inputenc}

\usepackage{microtype}

\usepackage{inconsolata}

\title{Selecting Shots for Demographic Fairness in Few-Shot Learning \\ with Large Language Models}
\author{Carlos Aguirre, Kuleen Sasse, Isabel Cachola \and Mark Dredze \\
        Center for Language and Speech Processing \\
        Johns Hopkins University \\
        \texttt{caguirre@cs.jhu.edu}}

\begin{document}
\include{commands}

\maketitle
\begin{abstract}
Recently, work in NLP has shifted to few-shot (in-context) learning, with large language models (LLMs) performing well across a range of tasks.
However, while fairness evaluations have become a standard for supervised methods, 
little is known about the fairness of LLMs as prediction systems. 
Further, common standard methods for fairness involve access to models weights or are applied during finetuning, which are not applicable in few-shot learning.
Do LLMs exhibit prediction biases when used for standard NLP tasks? 

In this work, we explore the effect of shots, which directly affect the performance of models, on the fairness of LLMs as NLP classification systems. 
We consider how different shot selection strategies, both existing and new demographically sensitive methods, affect model fairness across three standard fairness datasets. 
We discuss how future work can include LLM fairness evaluations.
\end{abstract}

\section{Introduction}

Historically, evaluation of machine learning systems concerned only overall performance; how well did a trained system do on a held-out test set. More recently, practitioners have realized that dataset level scores can mask uneven performance across different sets of data points \citep{barocas-hardt-narayanan}. 
This can be especially problematic when performance varies significantly between demographic groups, such as systems that do relatively worse on underrepresented and historically oppressed demographic groups \citep[e.g., ][]{zhang2020hurtful}. These systems are often called unfair or biased.

Fairness has implications for the quality of the user experience and system robustness, and can measure user experience in a manner not reflected by overall metrics. 
Additionally, fairness may have legal ramifications when AI regulations intersect with laws against discrimination \citep[e.g., ][]{kim2022race}. 
To address these disparities, researchers have developed methods for fairness that may be applied to training objectives, alignment after training, and evaluation metrics \citep{barocas-hardt-narayanan}.

A new approach to prediction relies on large language models (LLMs), in which an instance is accompanied by a prompt and a LLM relies on in-context learning to make a prediction \cite{Brown2020LanguageMA}. 
This type of learning, which requires no fine-tuning or other gradient updates, uses just a few examples at inference time as a ``prompt'' to guide inference on a final instance. 
Because in-context learning relies only on a few text examples during inference, the content of these examples can be very important for the quality of the emitted output \citep{dong2022survey}. 
While LLMs can do surprisingly well on various prediction tasks, models are measured 
once again on overall performance alone, not fairness, despite an understanding of the variable nature of LLM behavior \citep{chang2023language}. 
To date, little to no work has measured the fairness of LLMs as prediction systems, despite numerous studies showing inherent biases in the generations of LLMs \citep{stanczak2021survey}. 
Furthermore, traditional methods for addressing unfair models, whether pre-, in-, or post-training, are not applicable to LLMs as the data they're trained on is often proprietary, pre-training them is expensive, and many leading models are closed source.

Relying on the importance of the content of examples in few-shot learning, we study the fairness of LLMs as prediction systems considering how different demonstration selection methods affect the resulting social fairness of the model in classification tasks. 
Experiments with 7 popular models (Table~\ref{tab:models}) across 3 datasets find that LLMs are unfair predictors. 
We consider two types of demonstration selection methods to mitigate this unfairness: semantic and demographic-based, some novel and others from prior work.
We conduct an in-depth analysis of the performance and fairness of each demonstration selection method for each model.
While these selection methods can improve fairness, we see inconsistent improvements across datasets and models, suggesting future work to better understand how to achieve prediction fairness of LLMs.

\section{Data}

We consider three text classification datasets that include demographic information to 
evaluate the fairness of language models with regard to demographics: Bias in Bios \citep{10.1145/3287560.3287572}, Twitter Sentiment \citep{blodgett-etal-2016-demographic}, and HateXplain \citep{mathew2021hatexplain}.

\textbf{Bias in Bios} (demographics: gender) is a collection of English documents from CommonCrawl that contain biographies.
The task is to predict the occupation from the biography. 
\citet{10.1145/3287560.3287572} found gender bias present in models for this task.
Following \citet{kaneko-etal-2022-debiasing}, we measure gender bias by comparing the relative performance of models across biographies written about men and women. 
We select professions (labels) that had more than 1000 examples of biographies for each gender in the test set.\footnote{i.e. professions with at least 1000 men and 1000 women} 
This yields the following 8 labels: Attorney, Dentist, Journalist, Photographer, Physician, Professor, Psychologist, and Teacher. 
We randomly selected 500 for each gender from each profession to create a test set of 8,000 biographies. 
We then created a training set of 183,638 biographies by selecting all the biographies from the original train split with the professions listed above.

\textbf{Twitter Sentiment} (demographics: race) is a collection of English tweets where the task is to predict binary sentiment in a tweet. 
Tweets have also been annotated with a binary attribute corresponding to online text dialects: African-American English (AAE) or Standard American English (SAE), which has been previously correlated with parts-of-speech tagging performance difference in prior work \citep{blodgett-etal-2016-demographic}.
We use these text dialects as proxies for race and measure racial bias by comparing the relative performance of sentiment classification across the dialects, similar to \citet{shen-etal-2022-optimising}. 
To construct the dataset we follow \citet{han-etal-2022-fairlib}. We then select 40k and 2k random tweets from each combination of dialect and sentiment for train and test, creating a train set with 160k examples and test set of 8k.

\textbf{HateXplain} (demographics: race) is a collection of posts from Gab and Twitter annotated with toxicity and hate speech labels, as well as demographic labels for the target group of the hate speech.
While prior work has shown that there are performance differences for detecting hate speech for different target groups based on gender, religion, and race, we experiment only on race as it was the demographic characteristic with the reported highest disparities \citep{baldini-etal-2022-fairness}.
We remove Indigenous and Indian examples from our race demographics as they do not appear in all data splits. 
To construct the dataset, we followed a similar procedure to \citet{ye-etal-2021-crossfit}: we first reduced the space from multiclass to binary classification by combining the ``offensive'' and ``hatespeech'' labels to a singular ``toxic'' label while keeping the ``normal'' class the same. Because of HateXplain has multiple annotators per example for the labels and demographics, we take the majority label and the majority demographic. If there is not a majority in either, we discard the example.

\section{Methods}
We measure the effect of different demonstration selection methods on prediction fairness of LLMs. 
We hypothesize that, similar to how the choice of demonstrations has been shown to have an effect on performance, different methods of demonstration selection will affect social fairness of the model. 
This section describes the models evaluated, prompts, demonstration selection methods, and definitions of performance and fairness.
Overall, we conduct experiments in 36 setups (3 tasks, 12 models), using 6 demonstration selection strategies.

\subsection{Models}
We consider the fairness of several different LLMs, including open and closed source models. We consider both pretrained only (LLaMA \citep{touvron2023llama}, UL2 \cite{tay2023ul}, Llama2 \citep{touvron2023llama2}) and finetuned variants (Alpaca \citep{alpaca}, Flan-UL2 \cite{chung2022scaling}, Llama2-chat). 
We also consider two model sizes to observe the effects of size on fairness: LLaMA 7B and 65B, Alpaca 7B and 13B, and Llama2 13B and 70B. 
Finally, we consider two closed source models (\texttt{davinci-003}, \texttt{gpt-3.5-turbo}). \Cref{tab:models} shows the list of models tested in our experiments.

\begin{table}[t]
\resizebox{\columnwidth}{!}{
\centering
\begin{tabular}{cllc}
\toprule
\multicolumn{1}{l}{Access Type} & Model Name & Training Type     & Parameters \\ \cmidrule(lr){1-1} \cmidrule(lr){2-2} \cmidrule(lr){3-3} \cmidrule(lr){4-4}
\multirow{4}{*}{Open Source}    & LLaMA      & Pretrained        & 13B \& 65B  \\
                                & LLaMA2       & Pretrained \& chat & 13B \& 70B \\
                                  & Alpaca     & Instruction-tuned & 7B \& 13B  \\ 
                                  & UL2   & Pretrained        & 20B        \\ 
                                  & Flan-UL2        & Instruction-tuned & 20B        \\
                                  \midrule
\multirow{2}{*}{Closed Source}    & \texttt{davinci-003}       & Instruction-tuned        & 175B       \\ 
                                  & \texttt{gpt-3.5-turbo}    & Instruction-tuned\tablefootnote{\url{https://openai.com/blog/chatgpt}} & - \\
                                  \bottomrule 
\end{tabular}}
\caption{\label{tab:models}
The LLMs evaluated in this work.
}
\end{table}

\subsection{In-context Learning}

\begin{table*}[t]
\resizebox{\textwidth}{!}{
\begin{tabular}{c|c}
\toprule
\textbf{Dataset} & \textbf{Prompt Structure}\\
\hline
Bias in Bios & \texttt{<Bio>} \textbackslash n Occupations: \texttt{<List of Occupations>} \textbackslash nThe occupation of this person is \texttt{<label>}\\
TwitterAAE & Post:\texttt{<Tweet>}\textbackslash nQuestion: Is this post happy or sad? \textbackslash nAnswer: \texttt{<label>}\\
HateXplain & Post:\texttt{<Tweet>} \textbackslash nQuestion: Does this post contain offensive language?\textbackslash n Answer: \texttt{<label>}\\
\bottomrule
\end{tabular}}
\caption{\label{tab:prompts}
Prompt templates used in our experiments. For each example, $k=\{0,10\}$ demonstrations are constructed using the templates and prepended to the example which follows the same template but without the \texttt{<label>}.
}
\end{table*}
The focus of our experiments is on the effect that demonstrations have on fairness, however other aspects such as model hyperparameters and prompt structure may affect the performance of the model.
We conduct experiments varying temperature and choose the best ($1.0$) based on the results in \cref{apx:temperature-experiments}.
Further, we utilized existing prompts for each dataset where available. Otherwise, we adapted prompts from similar tasks. 
\Cref{tab:prompts} shows the prompt templates.
We choose the best prompt structures based on performance from past work, and leave exploration of the fairness effect of prompt structure to future work.

{\bf Bias in Bios}: We adapted the prompt from \citet{lin-etal-2022-shot} to include information about the labels. 
{\bf HateXplain}: We adopted the prompt from \citet{pmlr-v203-kocielnik23a}.
{\bf TwitterAAE}: Similar to Bias in Bios, we modified the prompt from \citet{min-etal-2022-rethinking} to include information about the labels. We prepended $k$ samples (shots) from the training set as demonstrations; each demonstration follows the same prompt format.
We evaluate models with zero-shot and 10-shot settings; we discontinued 5-shot evaluations after finding no meaningful differences in the results.

We note that it may be unrealistic to assume a large training set from which to draw demonstrations while also claiming a few-shot setting \citep{perez2021true}. If we indeed have hundreds or thousands of examples, train a model! Nevertheless, we evaluate in this setting to better understand the effects of demonstration selection on fairness. If one was going to annotate a small number of examples to include in a prompt, which type of examples should be included to maximize fairness? To answer this question, we rely on existing annotations (training sets) rather than creating our own.

\subsection{Demonstration Selection Strategies}
We evaluate existing demonstration selection methods for fairness: semantic similarity \citep{liu-etal-2022-makes, gao-etal-2021-making} and diversity \citep{zhang2022automatic}.
We also experiment with demographic-aware selection methods: sampling only \textit{within} the same demographic group and using a \textit{representative} sample.

\textbf{Zero-shot.} We contextualize the performance and fairness of shot selection methods by including zero-shot baselines, i.e. no added demonstrations.

\textbf{Random.}
We evaluate randomly selecting 10 demonstrations.
While this may not be optimal for performance \citep{liu-etal-2022-makes}, the fairness of this method is unknown.

\textbf{Similarity.} 
Demonstrations are selected based on the query instance. We select the $k=10$ most similar training examples as compared to the query instance. 
Similarity is measured based on the cosine distance of the SBERT \citep{reimers-gurevych-2019-sentence} embeddings, following \citet{gao-etal-2021-making}.\footnote{We use the \texttt{all-mpnet-base-v2} model which is the highest-performing sentence-embedding model at the time of writing.}

\textbf{Diversity.}
A single set of demonstrations is selected to include across all test instances to reflect a diversity of examples. 
Like Similarity selection, we obtain SBERT sentence embeddings and then use KMeans Clustering from the faiss library \citep{johnson2019billion} to produce $k=10$ clusters.
We selected the demonstrations with the vector closest to the centroid of each cluster \citep{zhang2022automatic}, in order to obtain samples that are semantically diverse.

\textbf{Within.}
We randomly select demonstrations that have the same demographic attribute as the test instance.
For example, in Bias in Bios, if the example is a biography of a woman, we randomly select biography demonstrations only from women.

\textbf{Representative.}
A single set of demonstrations is selected to include across all test instances to reflect a demographically representative set of instances. 
For example, in Bias in Bios, we randomly sample 5 biography demonstrations from women and 5 from men, obtaining a representative sample.

In addition to the demonstration selection methods, we experiment with appending the demographic category, e.g. race, sex, etc. (\texttt{demographic-attribute} prompting), to the prompt in each demonstration and the test example.
This is inspired by prior work that showed increased performance with demographically aware models \citep{hovy-2015-demographic}.

\subsection{Evaluation}
We obtain predictions by allowing each model to generate up to five tokens. Positive and negative labels are obtained by substring matching of the generated tokens.
Specifically, for Bias in Bios models, we allowed the term "lawyer" as correct for "attorney".
For performance, we report the macro-averaged F1 score of the model.

For the fairness evaluation, we use a modified 1-GAP metric originally introduced by \citet{10.1145/3287560.3287572}.
GAP is the difference in recall scores (TPR) between two demographic groups, also called \textit{equalized opportunity} \citep{hardt2016equality}.
We modified the definition to support multiple demographic groups by selecting the biggest recall difference across demographic groups, inspired by \citet{pmlr-v142-ghosh21a}.
We define the set of all demographics as $S$, $Y$ as the gold label, and $\hat{Y}$ as the prediction.  
\begin{equation*}
TPR_{s_i, y} = P\left(\hat{Y} = y \mid S  = s_i, Y = y\right)
\end{equation*}
\begin{equation*}
1-GAP = \min_{s_i, s_j \in S} 1 - (TPR_{s_i, y} - TPR_{s_j, y})
\end{equation*}

1-GAP gives us a relative metric, where models closest to 1 are the fairest.
However, to obtain a binary label for whether a model is fair, we obtain distributions of recall scores for each demographic by bootstrapping with 100 iterations.
We then perform a Krukal-Wallis (KW) one-way analysis of variance to test whether the recall score samples for each demographic belong to the same distribution (fair model.)

\subsection{Supervised and Other Baselines}
To contextualize the performance of the LLMs for these tasks, we compare the in-context models with a \textit{random classifier} baseline and BERT-based finetuned classification models with and without a fairness loss following \citet{foulds2020intersectional}.
The BERT-based classifiers are encoder+classification layer models that were end-to-end finetuned with the training data and hyperparameter tuned with the available dev sets.
The \textit{fairness} variants of BERT-based classifiers are finetuned with a true positive rate (TPR or recall-parity) using the demographics available per dataset \citep{foulds2020intersectional}.
We use BERT-style encoders \citep{devlin-etal-2019-bert} with vocabulary that match the dataset domain: RoBERTa for the Bias in Bios dataset \citep{DBLP:journals/corr/abs-1907-11692}
initialized with the \texttt{roberta-base} checkpoint,\footnote{\url{https://huggingface.co/roberta-base}}
and BERTweet for HateXplain and Twitter Sentiment \citep{nguyen-etal-2020-bertweet}, initialized with the \texttt{vinai/bertweet-base} checkpoint.\footnote{\url{https://huggingface.co/vinai/bertweet-base}}
For more model training details as well as the hyperparameter search space see \Cref{apx:finetune-details}.

\section{Results}

\Cref{tab:hatexplain_results} shows the results of the models on \textit{HateXplain} using the different demonstration selection methods, for all datasets see \cref{tab:all_selection_results} in \cref{apx:all-results}.
While the best performing LLMs are competitive compared to the supervised baselines, some settings perform below the random classifier baseline, as seen in \cref{tab:hatexplain_results} (UL2, LLaMA-13B\&65B, Alpaca-7B\&13B, and Llama2-13B\&70B).

For demographic fairness, we observe that the most fair models are often below random performance.
Since the ultimate goal of fairness is to maximize the utility of the models across all demographic groups (rather than none), we do not take into account fairness results from models that perform below a random classifier, these are shaded on \cref{tab:hatexplain_results}.
Comparing in-context models with BERT-based finetuned models, in-context models tend to be fairer but with a substantial loss in performance, with the most fair in-context model (zeroshot Llama2-70B-chat) performing $\approx 25$ F1 points lower than the fair BERT-based counterpart.
This is an extreme example of the fairness and accuracy trade-off, that is present in some of the LLMs we tested; fair models are fair because they perform poorly for all groups.

\subsection{Model Choice}

\begin{table*}[t]
\centering
\resizebox{.90\textwidth}{!}{
\begin{tabular}{rrlrlrlrlrlrl}
\toprule
                                        & \multicolumn{2}{l}{zeroshot} & \multicolumn{2}{l}{random} & \multicolumn{2}{l}{similarity} & \multicolumn{2}{l}{diversity} & \multicolumn{2}{l}{within} & \multicolumn{2}{l}{stratified} \\
\cmidrule(lr){2-3} \cmidrule(lr){4-5} \cmidrule(lr){6-7} \cmidrule(lr){8-9} \cmidrule(lr){10-11} \cmidrule(lr){12-13}
                                        & F1                        & 1-GAP                     & F1                        & 1-GAP                     & F1                        & 1-GAP                     & F1                        & 1-GAP                          & F1                        & 1-GAP                         & F1                           & 1-GAP                     \\
\cmidrule(lr){2-2} \cmidrule(lr){3-3} \cmidrule(lr){4-4} \cmidrule(lr){5-5} \cmidrule(lr){6-6} \cmidrule(lr){7-7} \cmidrule(lr){8-8} \cmidrule(lr){9-9} \cmidrule(lr){10-10} \cmidrule(lr){11-11} \cmidrule(lr){12-12} \cmidrule(lr){13-13}
\texttt{davinci-003}                    & 64.1                      & \textbf{84.7}             & \underline{\textbf{70.0}} & 74.0                      & 68.0                      & 78.0                      & 66.8                      & 69.6                           & 65.8                      & 82.6                          & 69.0                         & 79.5                      \\
\texttt{gpt-3.5-turbo}                  & 61.3                      & \textbf{85.6}             & \textbf{69.1}             & 80.5                      & 67.8                      & 73.8                      & 67.0                      & 80.8                           & 67.3                      & 82.1                          & 67.8                         & 78.6                      \\
UL2                                     & \textbf{53.5}             & \textbf{92.7}             & \cellcolor[gray]{0.8}44.3 & \cellcolor[gray]{0.8}99.1 & \cellcolor[gray]{0.8}44.3 & \cellcolor[gray]{0.8}96.7 & \cellcolor[gray]{0.8}44.4 & \cellcolor[gray]{0.8}100.0*    & \cellcolor[gray]{0.8}44.4 & \cellcolor[gray]{0.8}100.0*   & \cellcolor[gray]{0.8}44.3    & \cellcolor[gray]{0.8}96.8 \\
FLAN-UL2                                & 60.9                      & 71.0                      & 68.4                      & 83.8                      & 68.6                      & \textbf{85.6}             & 68.3                      & 83.5                           & 68.9                      & 82.3                          & \textbf{69.1}                & 82.6                      \\
LLaMA-13B                               & \cellcolor[gray]{0.8}22.3 & \cellcolor[gray]{0.8}77.5 & \cellcolor[gray]{0.8}31.3 & \cellcolor[gray]{0.8}69.1 & \textbf{48.5}            & \textbf{52.6}             & \cellcolor[gray]{0.8}23.5 & \cellcolor[gray]{0.8}75.7      & \cellcolor[gray]{0.8}36.0 & \cellcolor[gray]{0.8}48.7     & \cellcolor[gray]{0.8}32.0    & \cellcolor[gray]{0.8}78.2 \\
LLaMA-65B                               & \cellcolor[gray]{0.8}40.5 & \cellcolor[gray]{0.8}84.6 & \cellcolor[gray]{0.8}44.7 & \cellcolor[gray]{0.8}76.4 & \textbf{52.2}             & \textbf{79.6}             & 49.6                      & 60.7                           & 47.2                      & 71.3                          & 48.8                         & 68.7                      \\
Alpaca-7B                               & \cellcolor[gray]{0.8}28.7 & \cellcolor[gray]{0.8}87.9 & 48.8                      & 66.1                      & \textbf{52.2}             & 82.9                      & 45.6                      & 78.6                           & 45.7                      & 80.2                          & 48.9                         & \textbf{92.8}             \\
Alpaca-13B                              & \cellcolor[gray]{0.8}27.7 & \cellcolor[gray]{0.8}85.7 & \cellcolor[gray]{0.8}34.9 & \cellcolor[gray]{0.8}84.8 & \cellcolor[gray]{0.8}38.3 & \cellcolor[gray]{0.8}78.5 & \cellcolor[gray]{0.8}37.1 & \cellcolor[gray]{0.8}74.7      & \cellcolor[gray]{0.8}35.5 & \cellcolor[gray]{0.8}76.9     & \cellcolor[gray]{0.8}36.6    & \cellcolor[gray]{0.8}77.1 \\
LLaMA2-13B                              & \cellcolor[gray]{0.8}33.0 & \cellcolor[gray]{0.8}86.5 & 46.1                      & \textbf{94.6 }            & 47.1                      & 85.2                      & \textbf{47.1}             & 93.5                           & 46.0                      & 88.7                          & \cellcolor[gray]{0.8}43.9    & \cellcolor[gray]{0.8}92.6 \\
LLaMA2-13B-chat                         & \textbf{63.4}             & \textbf{93.5}             & 59.9                      & 71.1                      & 63.0                      & 65.2                      & 59.3                      & 49.2                           & 58.9                      & 93.3                          & 61.6                         & 81.5                      \\
LLaMA2-70B                              & \textbf{46.1}            & \textbf{90.9}             & \cellcolor[gray]{0.8}25.5 & \cellcolor[gray]{0.8}78.7 & 33.3                      & 77.2                      & \cellcolor[gray]{0.8}15.1 & \cellcolor[gray]{0.8}79.6      & \cellcolor[gray]{0.8}28.2 & \cellcolor[gray]{0.8}81.8     & \cellcolor[gray]{0.8}33.5    & \cellcolor[gray]{0.8}80.4 \\
LLaMA2-70B-chat                         & 48.5                      & \underline{\textbf{99.1}} & \textbf{51.9}             & 68.2                      & \cellcolor[gray]{0.8}42.4 & \cellcolor[gray]{0.8}74.6 & \cellcolor[gray]{0.8}31.7 & \cellcolor[gray]{0.8}82.2      & 46.4                      & 72.0                          & 51.1                         & 77.2                      \\
\cmidrule(lr){2-13}
\textit{random class.}  & 45.2  &  \\
\textit{BERTweet} & 72.7 & 40.0  \\
\textit{BERTweet Fair}  & 73.2 & 86.9 \\
\bottomrule
\end{tabular}}
    \caption{Macro-averaged F1 and 1-GAP on HateXplain dataset, \textbf{bold} is best per model, \underline{underlined} is best overall, asterisk (*) denotes absolute fairness,\footnotemark{} and \colorbox{gray!80}{shade} are results with F1 score below a random baseline.}
    \label{tab:hatexplain_results}
\end{table*}

When considering the overall performance of models across all our settings, it becomes clear that the choice of model matters both in terms of performance and fairness.
Flan-UL2, \texttt{davinci-003}, \texttt{gpt-3.5-turbo} and Llama2-13B-chat are the best-performing models across the three datasets.
Some models, e.g. Alpaca and UL2, have better than random performance in only one dataset.
In contrast, there is not a clear winner for fairness, with model fairness varying across all datasets.
However, the more drastic fairness differences are at the dataset level, where the fairness of all models in Twitter Sentiment ($>.9$ for all models) is much greater than, e.g. HateXplain.
When comparing fine-tuned vs pretrained variants of LLMs (FLAN-UL2 vs. UL2, LLaMA2 vs. LLama2-chat), finetuning seems to help in performance but have a varied effect on fairness. 

Overall, we find that model selection for fairness cannot be generalized across datasets.

\subsection{Performance and Fairness}

\begin{figure*}[t]
     \centering
     \begin{subfigure}[b]{0.3\textwidth}
         \centering
         \includegraphics[width=\textwidth]{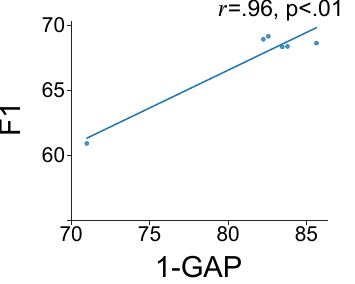}
         \caption{Flan-UL2}
         \label{fig:flan-ul2}
     \end{subfigure}
     \hfill
     \begin{subfigure}[b]{0.3\textwidth}
         \centering
         \includegraphics[width=\textwidth]{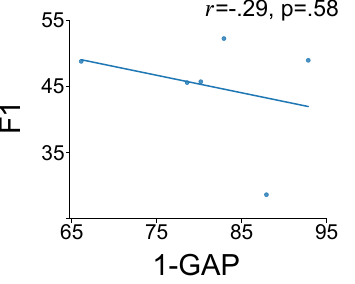}
         \caption{Alpaca-7B}
         \label{fig:gpt3}
     \end{subfigure}
     \hfill
     \begin{subfigure}[b]{0.3\textwidth}
         \centering
         \includegraphics[width=\textwidth]{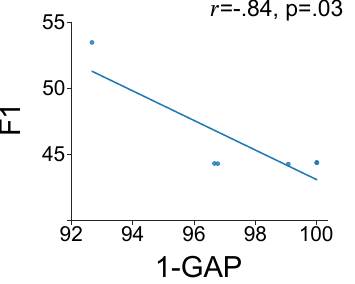}
         \caption{UL2 }
         \label{fig:ul2}
     \end{subfigure}
        \caption{F1 vs 1-GAP when varying demonstration selection methods for Flan-UL2, Alpaca-7B and UL2 in HateXplain dataset showing positive, no correlation and negative correlations respectively.}
        \label{fig:performance_v_fairness}
\end{figure*}

1-GAP (fairness) has an inherent connection with F1 (performance) since both include recall.
However, we can still have fair models at different ranges of accuracy. Many have postulated that there is a trade-off between fairness and performance; fairness comes at the expense of performance resulting in a negative correlation.
Much recently, \citet{islam2021can} showed this trade-off is not always present empirically; some methods obtain high performance and fairness.

Our experiments (perhaps distressingly) exhibit both positive and negative correlations for certain models across datasets.
\Cref{fig:performance_v_fairness} shows the 1-GAP vs F1 plots for three models, which have a positive (Flan-UL2), no (Alpaca-7B) and negative correlation (UL2) between performance and fairness. 
This erratic relationship underscores the need for explicit evaluation of fairness rather than relying on performance alone.

\begin{table}[t]
\resizebox{\columnwidth}{!}{
\begin{tabular}{rrlrlrl}
\toprule
           & \multicolumn{2}{c}{HateXplain} & \multicolumn{2}{c}{Bias in Bios} & \multicolumn{2}{c}{Twitter Sent.} \\
           \cmidrule(lr){2-3} \cmidrule(lr){4-5} \cmidrule(lr){6-7}
           & \multicolumn{1}{c}{F1} & \multicolumn{1}{c}{1-GAP} & \multicolumn{1}{c}{F1}    & \multicolumn{1}{c}{1-GAP} & \multicolumn{1}{c}{F1}    & \multicolumn{1}{c}{1-GAP} \\

           \cmidrule(lr){2-2} \cmidrule(lr){3-3} \cmidrule(lr){4-4} \cmidrule(lr){5-5} \cmidrule(lr){6-6} \cmidrule(lr){7-7}
zeroshot   & 45.8          & \textbf{86.6}  & 38.8           & \textbf{94.8}   & 39.6              & 96.6              \\
random     & 49.6          & 78.9           & 66.0           & 88.2            & 42.9              & 97.1              \\
similarity & \textbf{52.1} & 77.5           & 62.3           & 90.8            & \textbf{50.9}     & \textbf{97.7}     \\
diversity  & 46.3          & 77.3           & \textbf{66.1}  & 88.4            & 43.9              & 96.7              \\
within     & 49.2          & 80.0           & 65.8           & 89.2            & 43.2              & 94.6              \\
stratified & 50.6          & 82.2           & 64.4           & 89.6            & 43.0              & 96.8              \\
\bottomrule
\end{tabular}}
\caption{\label{tab:selection-strategy-best}
Mean F1 \& 1-GAP per selection strategy.
}
\end{table}

\subsection{Zero-shot Settings are Sometimes Better}
How important is adding demonstrations (few-shot) to prompts compared to leaving them out (zero-shot) for fairness? 
The effect is especially pronounced for UL2, LLaMA, and Alpaca, e.g. Alpaca-7B goes from unusable performance in zero-shot (2.3 F1) to decent in few-shot (82.1 F1) in Bias in Bios.
On the other hand, higher performing models (\texttt{davinci-003}, \texttt{gpt-3.5-turbo} and Flan-UL2) sometimes do better in the zero-shot setting; adding demonstrations hurts performance.
Nevertheless, on average across models, zero-shot settings were always outperformed by all demonstration selection methods (see \Cref{tab:selection-strategy-best}).

The relationship between demonstrations and fairness is more varied.
In general, when both fairness and performance in zeroshot settings are high, adding demonstrations does not help and can even harm fairness.
However, in average across models, zeroshot settings are generally more fair than other demonstration selection methods closely followed by \textit{similarity}.
While adding demonstrations helps performance, the effect on fairness is unpredictable. This again underscores the importance of evaluating prediction fairness of LLMs.

\subsection{Which Demonstrations To Add}
Adding demonstrations (Random vs. Zero-shot) usually improves model performance ($\sim$70\% of the time), but often made model fairness worse ($\sim$60\% of the time was worse).
Care in demonstration selection is needed to ensure fairness.

For \textit{similarity} and \textit{diversity} selection methods: \textit{similarity} selection helps performance on average across datasets compared to random selection and zero-shot (\cref{tab:selection-strategy-best}.)
This same is generally true for fairness, but still less fair than zeroshot.
In contrast, \textit{Diversity} selection has less consistent behavior, where it helps LLaMA-65B and Flan-UL2, but hurts every other model.
The fairness scores also fluctuate and vary by data and model.
We also observe fluctuations with demographic-based demonstration selection strategies, albeit with less success overall.
Perhaps surprisingly, selecting demonstrations from \textit{within} the same demographic was the least favored settings in both performance and fairness across models and datasets. We expected choosing data of the same type would help fairness; it did not.
A \textit{representative} selection of demonstrations had more success than \textit{within} in both performance and fairness.

While \textit{similarity} selection was the most helpful in both performance and fairness, we would hope that there exists a single demonstration selection strategy that consistently improves performance and fairness. Unfortunately, this was not the case.

\subsection{Including Demographic Attributes}
\label{sec:results-protected-attributes}

\begin{figure}[t]
 \centering
 \includegraphics[]{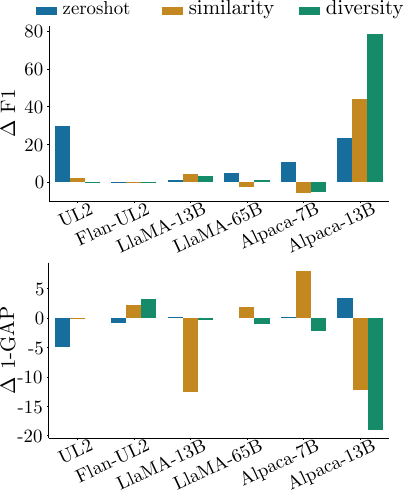}
 \caption{$\Delta$ F1 and $\Delta$ 1-GAP when including demographic attributes in prompt on Bias in Bios. 
 }
 \label{fig:demographic_prompting_results}
\end{figure}
\footnotetext{The recall scores from bootstrap samples (100) across demographics belong to the same distribution.}
Perhaps telling the model demographic information can reduce bias in the output.
\Cref{fig:demographic_prompting_results} shows the results of including demographic attributes with the demonstrations to open source models in the Bias in Bios dataset (all datasets are shown in \Cref{tab:protected-attribute-results}). While adding demographic attributes helps in terms of performance, benefits appear to be model specific.
For LLaMA and Alpaca, some settings have improved performance, but overall a mixed effect on fairness, e.g. for Alpaca-13B with demonstrations selected with \textit{diversity} the performance increased from $2$ F1 to $80$ by simply adding the demographic attributes but, at the same time, reduced from perfect fairness ($100$) to $81$ (\Cref{fig:demographic_prompting_results}.)
Adding demographic attributes affected the performance and fairness of Flan-UL2 models to a lesser effect.
For these models, there was a general trade-off between increasing performance but decreasing fairness, and vice-versa.

Overall, adding demographic attributes seems to help LLaMA and Alpaca models the most in performance, perhaps because more information is provided, but the effect on fairness is mixed.

\subsection{Other Selection Methods}
\label{sec:results-cross-methods}

Since \textit{similarity} and \textit{diversity} selection were more successful than demographic-based selection, we experimented with combining these and the \textit{within} method.
We test \textit{within}+\textit{similarity}, demonstrations that are most similar within the same demographic group, and \textit{within}+\textit{diversity}, demonstrations that are most diverse within the same demographic.

\Cref{fig:cross_methods_results} show results for Bias in Bios and \Cref{tab:combining-selections-results} for all datasets.
Unfortunately, combining \textit{within} and \textit{similarity} methods often drastically {\bf decreases} model performance, but sometimes increases fairness (Flan-UL2.)
This is interesting as these are the most similar methods, with $\sim80\%$ of demonstrations selected by \textit{similarity} being within the same demographic.
Despite these similarities, we see that semantic \textit{similarity} is generally more important than demographic similarity for both performance and fairness, and combining these two actually hinders the performance of the models.

On the other hand, combining \textit{within} and \textit{diversity} selection methods often helps in both performance and fairness!
Contextualizing these results with the previous subsections, a rule-of-thumb is to select semantically diverse demonstrations within the same demographic group, or semantically similar demonstrations across all demographics.

While semantic similarity was not always the best performing, it provides the best performance and fairness trade-off between the demonstration selection methods.

\section{Related Work}
\textbf{In-Context Learning.}
Large Language Models are effective in a large number of classification and generative tasks~\cite{Devlin2019BERTPO, Radford2019LanguageMA, Liu2019RoBERTaAR, Lewis2019BARTDS}. While finetuning a pretrained model is a popular paradigm~\cite{Devlin2019BERTPO}, finetuning large models can be cost-prohibitive because of the compute required to do so. Furthermore, finetuning requires additional task-specific labeled data, which can also be prohibitively expensive to collect. \citet{Brown2020LanguageMA} evaluated \textit{in-context learning}, or few-shot learning, for LLMs, a learning paradigm in which the model is given a few examples, or demonstrations, of a task and is then asked to complete the final example. In-context learning has shown impressive results in a variety of tasks, including question answering, translation, and natural language inference~\cite{Brown2020LanguageMA}. 

Work on in-context learning has focused on writing better prompts~\cite{Wei2022ChainOT, Min2021NoisyCL, Holtzman2021SurfaceFC, Zhao2021CalibrateBU}, choosing better demonstrations~\cite{Liu2021WhatMG, Rubin2021LearningTR}, and training with an in-context learning objective~\cite{Min2021MetaICLLT, Chen2021MetalearningVL}. There have also been explorations of the sensitivities of in-context learning, such as the format of the prompts \cite{Gao2021MakingPL, Jiang2019HowCW} or the order of the demonstrations~\cite{Lu2021FantasticallyOP}. However, prior work has not studied the effect of demonstration choice on social fairness, only on overall performance \citep{dong2022survey}. 
Other work, like \citet{ma2023fairness} has evaluated the \textit{label fairness}, i.e. performance differences across different labels or classes in a multi-class prediction setting, of LLMs in in-context learning by creating a system that chooses prompts to create a "fair" demonstration. Similar to our work, they focused on shot or demonstration choice and found that shot selection matters for performance. 
Thus, given the minimal amount of data used for in-context learning, we suspect that the choice of demonstrations has an effect on the social fairness of the model's output.

\textbf{Social Fairness with Large Language Models.}
Work that identifies and measures the biases of language models have classified these harms in two general categories: \textit{allocation} and \textit{representation} harm \citep{stanczak2021survey}.
Representational harms happen when harmful concepts or relations are associated with demographic groups by a model; in language models these are often measured via token embeddings and model parameters with fill-in the blank, or complete the sentence templates \citep[e.g., ][]{nadeem-etal-2021-stereoset,nangia-etal-2020-crows}.
Most bias studies in NLP have focused on representational harms: many studies have demonstrated how generations from LLMs exhibit bias towards specific groups, or generate text that can be considered offensive, harmful or toxic~\citep{dodge-etal-2021-documenting, 10.1145/3287560.3287572, 10.1145/3442188.3445922, nadeem-etal-2021-stereoset}, generations from LLMs are more likely to generative negative sentiment for refugees, disabled people, AAVE sentences, nonbinary, muslim and women \citep{magee2021intersectional, groenwold-etal-2020-investigating, sheng-etal-2019-woman}. 
To understand the underlying bias source in the behavior of these models, researchers have created methods for evaluating the generations of LLMs under different conditions, like size and training procedure \citep{baldini-etal-2022-fairness, tal-etal-2022-fewer, de-vassimon-manela-etal-2021-stereotype, nangia-etal-2020-crows}.

On the other hand, allocational harms are reflected on performance differences on data associated with different demographic groups \citep{stanczak2021survey}, also known as fairness.
Little work has focused on allocation harms from in-context learning in LLMs for classification settings.
\citet{salewski2023context} found that impersonating roles improves performance for in-context learning on LLMs: impersonating an expert in a task can improve performance of the model for that task; however, these impersonations can also reveal biases in models by finding disparate performances from impersonating different roles, e.g. better performance when impersonating men than women.
Perhaps the most related work is \citet{zhang2022fairness}, who investigates fairness re-programming techniques for models that cannot be re-trained or finetuned, e.g. in-context learning LLMs.
They append token perturbations to the prompt, \textit{fairness triggers}, that are learned from a helper model.
They show that by appending false pseudo-demographic information, they can decrease performance differences across demographic groups.
We, instead, focus on investigating the role of choice of demonstrations or shots in the performance differences of LLMs on in-context learning settings.

\section{Conclusion}
Significant work has gone into evaluating different demonstration selection strategies in the performance of LLMs as prediction systems. This paper represents one of the first studies to consider the fairness of these systems. Our study considers 7 widely used family of models (\Cref{tab:models}), three datasets, and multiple demonstration selection methods.

We find that model selection for fairness cannot be generalized across datasets.
While Flan-UL2 is among the best-performing and fairest models, there is unfortunately no clear winner across all three datasets and they still underperform compared to supervised baselines often with a more drastic fairness vs performance trade-off.
In terms of shot selection strategies, while adding demonstrations (with the best selection method) generally yields higher performing models (compared to \textit{zero-shot}), it does not consistently yield fairer models.
While we cannot say that a single selection method performs the best across all datasets and models, or even always helps improve fairness, our experiments suggest that, on average, \textit{similarity} is the better option.

Where do these results leave us? First, fairness {\bf must} be evaluated alongside task performance when developing prompts, selection strategies, and models. We cannot assume any relationship between fairness and performance. Second, we need to better understand {\bf why} LLMs are unfair in their predictions. While significant work has examined fairness in supervised training objectives \citep{delobelle2021measuring}, and other work demonstrates bias in LLM generations \citep{chang2023language}, we need work that intersects these two. Third, how can we determine {\bf when} a LLM is being unfair? Work examining confidence in LLM predictions \citep[e.g.,][]{portillo-etal-2023-strength} can help automatically determine the accuracy of the system. Can we develop similar metrics for fairness? This would be especially helpful in cases where we do not have demographically labeled data. Finally, there is now a large focus on fine-tuning LLMs (e.g. RLHF \citep{ouyang2022training}, FLAN \citep{chung2022scaling}). The goal of these methods has been better instruction following and improved accuracy on prediction tasks, but our results suggest they do not always make models fairer. How can we include fairness objectives in this training process?

\section*{Limitations}
We work with LLMs that are expensive to run (large GPUs to run big open source models) 
or costly to access (cost of APIs). This limits our ability to fully explore all possible methods.
For example, OpenAI API costs precluded our use of close-source models in some experiments \Cref{sec:results-protected-attributes,sec:results-cross-methods}.
Furthermore, our closed-source model evaluations may not be reproducible as we do not have control over updates to the underlying models and the model outputs are known to be inconsistent \citep{ye2023assessing}.

While we consider 8 models, there are now many different LLMs available for evaluation, with several released concurrent with this study, e.g. Falcon \citep{falcon40b} and Vicuna \citep{vicuna2023}. We cannot evaluate all models, but our results suggest that the fairness of these models will also be highly varied. 
Additionally, other aspects of in-context learning may also affect the fairness of LLMs that we did not study, e.g. demonstration ordering \citep{lu-etal-2022-fantastically} and prompt formatting \citep{wang2022self}.

\section*{Ethics Statement}
We study the fairness of language models for three tasks: occupation classification, sentiment analysis, and hate speech detection.
Occupation classification has direct applications in the automation of hiring procedures, which have been historically biased along many more demographic attributes than what we consider, e.g. age, disabilities, race, ethnicity, sexual orientation, and veteran status.
The same is true of the other datasets in this paper.
Additionally, often these inequities intersect across these social groups, further increasing the impact of applications that use these models outside of an academic environment.
Because we were limited by the currently available datasets and the coverage they have on demographic attributes, we acknowledge that fairness as is discussed in this paper will not translate to social fairness in the wild without first considering all of these biases.

\section*{Acknowledgements}
This work was carried out at the Advanced Research Computing at Hopkins (ARCH) core facility  (rockfish.jhu.edu), which is supported by the National Science Foundation (NSF) grant number OAC1920103.

\bibliography{anthology,custom}
\bibliographystyle{acl_natbib}

\newpage

\appendix

\section{All Results}
\label{apx:all-results}

\begin{figure}[t]
 \centering
 \includegraphics[width=\columnwidth]{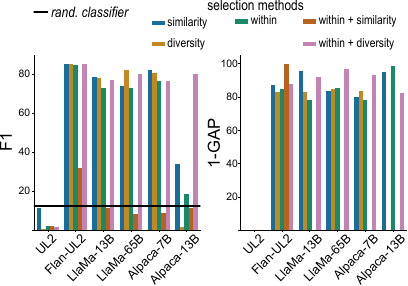}
 \caption{Performance (F1) and fairness (1-GAP) of combining \textit{within} with semantic-based methods across models in the Bias in Bios dataset. For 1-GAP graph we show models with $>$ rand. classifier performance.}
 \label{fig:cross_methods_results}
\end{figure}

Here we present \Cref{tab:all_selection_results}, containing the results in performance (macro-averaged F1) and fairness (1-GAP) for all models, selection methods and datasets. We also show the performance of the models adding demographic attributes to the demonstrations and prompt in \Cref{tab:protected-attribute-results}. And finally, we show the performance and fairness of the models when combining semantic and demographic based selection methods in \Cref{tab:combining-selections-results} and \Cref{fig:cross_methods_results}.

\begin{table*}[t]
\centering
\resizebox{.90\textwidth}{!}{
\begin{tabular}{rrlrlrlrlrlrl}
\toprule
\multicolumn{13}{c}{\textbf{HateXplain race}} \\
                                        & \multicolumn{2}{l}{zeroshot} & \multicolumn{2}{l}{random} & \multicolumn{2}{l}{similarity} & \multicolumn{2}{l}{diversity} & \multicolumn{2}{l}{within} & \multicolumn{2}{l}{stratified} \\
\cmidrule(lr){2-3} \cmidrule(lr){4-5} \cmidrule(lr){6-7} \cmidrule(lr){8-9} \cmidrule(lr){10-11} \cmidrule(lr){12-13}
                                        & F1                        & 1-GAP                     & F1                        & 1-GAP                     & F1                        & 1-GAP                     & F1                        & 1-GAP                          & F1                        & 1-GAP                         & F1                           & 1-GAP                     \\
\cmidrule(lr){2-2} \cmidrule(lr){3-3} \cmidrule(lr){4-4} \cmidrule(lr){5-5} \cmidrule(lr){6-6} \cmidrule(lr){7-7} \cmidrule(lr){8-8} \cmidrule(lr){9-9} \cmidrule(lr){10-10} \cmidrule(lr){11-11} \cmidrule(lr){12-12} \cmidrule(lr){13-13}
\texttt{davinci-003}                    & 64.1                      & \textbf{84.7}             & \underline{\textbf{70.0}} & 74.0                      & 68.0                      & 78.0                      & 66.8                      & 69.6                           & 65.8                      & 82.6                          & 69.0                         & 79.5                      \\
\texttt{gpt-3.5-turbo}                  & 61.3                      & \textbf{85.6}             & \textbf{69.1}             & 80.5                      & 67.8                      & 73.8                      & 67.0                      & 80.8                           & 67.3                      & 82.1                          & 67.8                         & 78.6                      \\
\bottomrule
\end{tabular}}
    \caption{Macro-averaged F1 score and 1-GAP of all models and demonstration selection methods for all of the three datasets. \textbf{Bold} is best per model$\times$dataset and \underline{underlined} is best per dataset (above a random baseline). Asterisk (*) denotes no significant difference in recall scores performing a Kruskal-Wallis test with 100 bootstrap iterations. We \colorbox{gray!80}{shade} results that have an F1 score below a random baseline.}
    \label{tab:all_selection_results}
\end{table*}

\begin{table*}
\centering
\resizebox{.85\textwidth}{!}{
\begin{tabular}{ll@{\hskip -15mm}lllllllllllll}
\toprule
                             &           &               & \multicolumn{4}{c}{HateXplain race}                & \multicolumn{4}{c}{Bias in Bios}                   & \multicolumn{4}{c}{Twitter Sentiment}              \\
                           \cmidrule(lr){4-7} \cmidrule(lr){8-11} \cmidrule(lr){12-15}
                        & & & \multicolumn{2}{c}{F1 ($\Delta$)} & \multicolumn{2}{c}{1-GAP ($\Delta$)} & \multicolumn{2}{c}{F1 ($\Delta$)} & \multicolumn{2}{c}{1-GAP ($\Delta$)} & \multicolumn{2}{c}{F1 ($\Delta$)} & \multicolumn{2}{c}{1-GAP ($\Delta$)} \\
 \cmidrule(lr){4-5} \cmidrule(lr){6-7} \cmidrule(lr){8-9} \cmidrule(lr){10-11} \cmidrule(lr){12-13} \cmidrule(lr){14-15}
\textit{baseline} & \textit{random class.} & & 61.3 &  & & & 12.5 &  & & & 50.0 & \\
\midrule
model                        & \multicolumn{2}{l}{selection method}\\
\cmidrule(lr){1-1} \cmidrule(lr){2-3}
\multirow{12}{*}{UL2}        & zero-shot   &              & \cellcolor[gray]{0.8}\textbf{53.5} &                          & \cellcolor[gray]{0.8}92.7 &                                          & 19.2 &                                           & 99.6 &                                           & 58.1 &                                           & 98.6 &                                  \\
                              & & $+$demographic attributes &\cellcolor[gray]{0.8}45.9 & \cellcolor[HTML]{F9CB9C}(-7.6) & \cellcolor[gray]{0.8}\textbf{100} & \cellcolor[HTML]{9FC5E8}(7.3)  & \textbf{48.7} & \cellcolor[HTML]{9FC5E8}(29.5)  & 94.6 & \cellcolor[HTML]{F9CB9C}(-5.0)         & 61.1 & \cellcolor[HTML]{9FC5E8}(3.0)           & 78.8 & \cellcolor[HTML]{F9CB9C}(-19.8) \\
\bottomrule

\end{tabular}}
\caption{\label{tab:protected-attribute-results}
Performance of open source models across datasets when adding demographic attributes to the demonstrations and prompt. Results without demographic attributes are shown as comparison, as well as a difference between them. \textbf{Bold} is best per model$\times$dataset and \underline{underlined} is best per dataset (above a random baseline). We \colorbox{gray!80}{shade} results that have an F1 score below a random baseline.
}
\end{table*}

\begin{table*}
\centering
\resizebox{.75\textwidth}{!}{
\begin{tabular}{ll@{\hskip -10mm}lllllll}
\toprule
                          &                         &            & \multicolumn{2}{c}{HateXplain race} & \multicolumn{2}{c}{Bias in Bios} & \multicolumn{2}{c}{Twitter Sentiment} \\
                          \cmidrule(lr){4-5} \cmidrule(lr){6-7} \cmidrule(lr){8-9}

model                     & \multicolumn{2}{c}{selection method}   & \multicolumn{1}{c}{F1}               & 1-GAP            & \multicolumn{1}{c}{F1}              & 1-GAP          & \multicolumn{1}{c}{F1}                & 1-GAP             \\
\cmidrule(lr){1-1}  \cmidrule(lr){2-3} \cmidrule(lr){4-4} \cmidrule(lr){5-5} \cmidrule(lr){6-6} \cmidrule(lr){7-7} \cmidrule(lr){8-8} \cmidrule(lr){9-9}
\multirow{8}{*}{UL2}        & zero-shot           &                  & \cellcolor[gray]{0.8}\textbf{53.5} & \cellcolor[gray]{0.8}92.7          & \textbf{19.2}                & 99.6                                  & 58.1                                  & 98.6                                  \\
                             & random            &                  & \cellcolor[gray]{0.8}44.3          & \cellcolor[gray]{0.8}99.1          & \cellcolor[gray]{0.8}2.5 & \cellcolor[gray]{0.8}\textbf{100} & \cellcolor[gray]{0.8}48.2          & \cellcolor[gray]{0.8}92.6          \\
                             & similarity        &                  & \cellcolor[gray]{0.8}44.3          & \cellcolor[gray]{0.8}96.7          & \cellcolor[gray]{0.8}11.5 & \cellcolor[gray]{0.8}\textbf{100} & \textbf{65.0}                         & 99.9                                  \\
                             & diversity         &                  & \cellcolor[gray]{0.8}44.4          & \cellcolor[gray]{0.8}\textbf{100} & \cellcolor[gray]{0.8}0.9 & \cellcolor[gray]{0.8}\textbf{100} & \cellcolor[gray]{0.8}33.5          & \cellcolor[gray]{0.8}\textbf{100} \\
\bottomrule
\end{tabular}}
\caption{\label{tab:combining-selections-results}
Performance of open source models across datasets for demonstration selection methods that select based on semantic similarity within the same demographic category (\textit{within + similarity}) and semantic diversity within the same demographic (\textit{within + diversity}). We show results for other selection methods for context. \textbf{Bold} is best per model$\times$dataset and \underline{underlined} is best per dataset (above a random classifier baseline). We \colorbox{gray!80}{shade} results that have an F1 score below a random class. baseline.
}
\end{table*}

\section{BERT-based fine-tuning details}
\label{apx:finetune-details}
We use BERT-style encoders \citep{devlin-etal-2019-bert} with a vocabulary match the dataset domain: RoBERTa for the Bias in Bios dataset \citep{DBLP:journals/corr/abs-1907-11692}
initialized with the \texttt{roberta-base} checkpoint,\footnote{\url{https://huggingface.co/roberta-base}}
and BERTweet for HateXplain and Twitter Sentiment \citep{nguyen-etal-2020-bertweet}, initialized with the \texttt{vinai/bertweet-base} checkpoint.\footnote{\url{https://huggingface.co/vinai/bertweet-base}}
We add a separate linear classification head for each task, with a Softmax output function to allow for multi-class classification (Bias in Bios) or a Sigmoid output function for binary classification (HateXplain and Twitter Sentiment.) 
The document representation for the classification head is a mean-pooled aggregation across all subword representations of the document taken at the top layer of the network..
Models were trained on Nvidia A100 GPUs, using \texttt{jiant} \citep{phang2020jiant}, a multi-task wrapper library.

In addition to a typical finetuning model, we also provide a finetuned model with an added fairness loss, to compare with a model that adds fairness to the objective.
We utilize equalized opportunity, also known as GAP, as our fairness definition, which is the compliment of 1-GAP, the fairness definition in the main paper. 
We use $\epsilon$-Differential Equalized Opportunity ($\epsilon$-DEO), a variant of $\epsilon$-DF \citep{foulds2020intersectional}, that applies the equalized opportunity objective, to ensure that the recall rates are equal across demographic groups \citep{barocas-hardt-narayanan} and that is learnable and differentiable. 

Formally, let $s_1, ... , s_p$ be discrete-valued demographic attributes, $z = s_1 \times s_2 \times ... \times s_p$. A model $M(X)$ satisfies $\epsilon$-DEO with respect to $z$ if for all $x$, $\hat{y} \in \textrm{Range}(M)$ and $y \in \textrm{Range}(M)$,
\begin{equation}
    e^{-\epsilon} \leq \frac{Pr(M(x) = 1 | s_i, y=1)}{Pr(M(x) = 1 | s_j, y=1)} \leq e^{\epsilon},
\end{equation}
for all $(s_i, s_j) \in z \times z$ where $Pr(s_i) > 0$, $Pr(s_j) > 0$; smaller $\epsilon$ is better, with $\epsilon$ = 0 for perfect fairness.
Perfect fairness results from a classifier with the same recall rates across groups of demographic attributes.

The standard approach to incorporating fairness metrics into learning objectives uses an additive term. For example, for a deep neural network classifier $M(X)$ with parameters $\theta$, we obtain the following,

\begin{multline}
    \min_{\theta} f(X; \theta) \overset{\Delta}{=} \frac{1}{N} \sum_{i=1}^{N} \mathcal{L}(x_i; \theta) \\
    + \lambda[\max(0, \epsilon(X;\theta) - \epsilon_t)]
\end{multline}


where $\epsilon(X;\theta)$ is the $\epsilon$-DEO measure for the classifier, $\epsilon_t$ is the desired base fairness (in our experiments 0), and $\lambda$ is a hyper-parameter that trades between prediction loss and fairness \citep{foulds2020intersectional}.
Since the fairness term is differentiable, the model can be trained using stochastic gradient descent on the objective via backpropagation and automatic differentiation.
A \textit{burn-in} period and stochastic approximation-based update are adopted following \citet{foulds2020intersectional}.

To obtain the best performing model, we use a grid search for each task, with a learning rate$=[1e^{-4}, 1e^{-5}, 1e^{-6}]$ with Adam optimizer \citep{kingma2014adam}, batch size$=[16, 32, 48]$, warmup$=[.1, .05, .005]$, epsilon$=[1e-7, 1e-8, 1e-9]$, \textit{burn-in}$=[.5, 1]$, $\lambda=[.01, .1]$ and $\rho=[.9, .1, .01]$. 
We select the best performing model on development data and report test data results.

\section{Hyperparameter Experiments}
\label{apx:temperature-experiments}

When considering the performance of LLMs for classification it may be important finetune the hyperparameters for generation.
In this section, we report the result of experiments when varying the temperature parameter across datasets.
Since we evaluate on 12 models across 3 datasets and 6 demonstration selection methods (total of 216 settings), varying the temperature for all settings is not practical.
Thus, we select the best performing open-source model, FLAN-UL2 for this experiment.

\begin{figure*}[t]
     \centering
     \includegraphics[width=\textwidth]{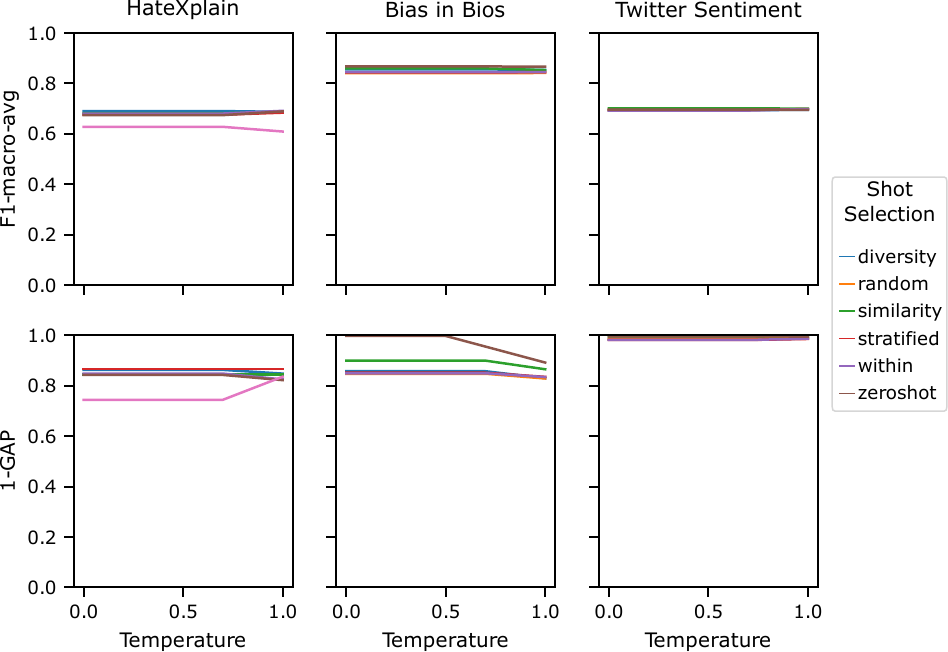}
        \caption{Results of varying temperature across datasets for Flan-UL2. No meaningful difference found.}
        \label{fig:temperature_experiments}
\end{figure*}

\Cref{fig:temperature_experiments} shows the results for performance (F1) and fairness (1-GAP) for FLAN-UL2 across all three datasets.
We observe little difference when varying temperature in the classification performance and the fairness of the model across demonstration selection strategies.

\end{document}

%% file: commands.tex
\newcommand\isabel[1]{{\color{purple}\{\textit{#1}\}$_{Isabel}$}}

\newcommand\todoit[1]{{\color{red}\{TODO: \textit{#1}\}}}
\newcommand\todo{{\color{red}{TODO}}\xspace}
\newcommand\todocite{{\color{red}{CITE}}\xspace}
\newcommand{\red}[1]{\textcolor{red}{#1}} 
\newcommand{\blue}[1]{\textcolor{blue}{#1}} 

%% file: main.bbl
\begin{thebibliography}{74}
\expandafter\ifx\csname natexlab\endcsname\relax\def\natexlab#1{#1}\fi

\bibitem[{Almazrouei et~al.(2023)Almazrouei, Alobeidli, Alshamsi, Cappelli,
  Cojocaru, Debbah, Goffinet, Heslow, Launay, Malartic, Noune, Pannier, and
  Penedo}]{falcon40b}
Ebtesam Almazrouei, Hamza Alobeidli, Abdulaziz Alshamsi, Alessandro Cappelli,
  Ruxandra Cojocaru, Merouane Debbah, Etienne Goffinet, Daniel Heslow, Julien
  Launay, Quentin Malartic, Badreddine Noune, Baptiste Pannier, and Guilherme
  Penedo. 2023.
\newblock {Falcon-40B}: an open large language model with state-of-the-art
  performance.

\bibitem[{Baldini et~al.(2022)Baldini, Wei, Natesan~Ramamurthy, Singh, and
  Yurochkin}]{baldini-etal-2022-fairness}
Ioana Baldini, Dennis Wei, Karthikeyan Natesan~Ramamurthy, Moninder Singh, and
  Mikhail Yurochkin. 2022.
\newblock \href {https://doi.org/10.18653/v1/2022.findings-acl.176} {Your
  fairness may vary: Pretrained language model fairness in toxic text
  classification}.
\newblock In \emph{Findings of the Association for Computational Linguistics:
  ACL 2022}, pages 2245--2262, Dublin, Ireland. Association for Computational
  Linguistics.

\bibitem[{Barocas et~al.(2019)Barocas, Hardt, and
  Narayanan}]{barocas-hardt-narayanan}
Solon Barocas, Moritz Hardt, and Arvind Narayanan. 2019.
\newblock \emph{Fairness and Machine Learning: Limitations and Opportunities}.
\newblock fairmlbook.org.
\newblock \url{http://www.fairmlbook.org}.

\bibitem[{Bender et~al.(2021)Bender, Gebru, McMillan-Major, and
  Shmitchell}]{10.1145/3442188.3445922}
Emily~M. Bender, Timnit Gebru, Angelina McMillan-Major, and Shmargaret
  Shmitchell. 2021.
\newblock \href {https://doi.org/10.1145/3442188.3445922} {On the dangers of
  stochastic parrots: Can language models be too big?}
\newblock In \emph{Proceedings of the 2021 ACM Conference on Fairness,
  Accountability, and Transparency}, FAccT '21, page 610–623, New York, NY,
  USA. Association for Computing Machinery.

\bibitem[{Blodgett et~al.(2016)Blodgett, Green, and
  O{'}Connor}]{blodgett-etal-2016-demographic}
Su~Lin Blodgett, Lisa Green, and Brendan O{'}Connor. 2016.
\newblock \href {https://doi.org/10.18653/v1/D16-1120} {Demographic dialectal
  variation in social media: A case study of {A}frican-{A}merican {E}nglish}.
\newblock In \emph{Proceedings of the 2016 Conference on Empirical Methods in
  Natural Language Processing}, pages 1119--1130, Austin, Texas. Association
  for Computational Linguistics.

\bibitem[{Brown et~al.(2020)Brown, Mann, Ryder, Subbiah, Kaplan, Dhariwal,
  Neelakantan, Shyam, Sastry, Askell, Agarwal, Herbert-Voss, Krueger, Henighan,
  Child, Ramesh, Ziegler, Wu, Winter, Hesse, Chen, Sigler, Litwin, Gray, Chess,
  Clark, Berner, McCandlish, Radford, Sutskever, and
  Amodei}]{Brown2020LanguageMA}
Tom~B. Brown, Benjamin Mann, Nick Ryder, Melanie Subbiah, Jared Kaplan,
  Prafulla Dhariwal, Arvind Neelakantan, Pranav Shyam, Girish Sastry, Amanda
  Askell, Sandhini Agarwal, Ariel Herbert-Voss, Gretchen Krueger, T.~J.
  Henighan, Rewon Child, Aditya Ramesh, Daniel~M. Ziegler, Jeff Wu, Clemens
  Winter, Christopher Hesse, Mark Chen, Eric Sigler, Mateusz Litwin, Scott
  Gray, Benjamin Chess, Jack Clark, Christopher Berner, Sam McCandlish, Alec
  Radford, Ilya Sutskever, and Dario Amodei. 2020.
\newblock Language models are few-shot learners.
\newblock \emph{ArXiv}, abs/2005.14165.

\bibitem[{Chang and Bergen(2023)}]{chang2023language}
Tyler~A Chang and Benjamin~K Bergen. 2023.
\newblock Language model behavior: A comprehensive survey.
\newblock \emph{arXiv preprint arXiv:2303.11504}.

\bibitem[{Chen et~al.(2021)Chen, Zhong, Zha, Karypis, and
  He}]{Chen2021MetalearningVL}
Yanda Chen, Ruiqi Zhong, Sheng Zha, George Karypis, and He~He. 2021.
\newblock Meta-learning via language model in-context tuning.
\newblock \emph{ArXiv}, abs/2110.07814.

\bibitem[{Chiang et~al.(2023)Chiang, Li, Lin, Sheng, Wu, Zhang, Zheng, Zhuang,
  Zhuang, Gonzalez, Stoica, and Xing}]{vicuna2023}
Wei-Lin Chiang, Zhuohan Li, Zi~Lin, Ying Sheng, Zhanghao Wu, Hao Zhang, Lianmin
  Zheng, Siyuan Zhuang, Yonghao Zhuang, Joseph~E. Gonzalez, Ion Stoica, and
  Eric~P. Xing. 2023.
\newblock \href {https://lmsys.org/blog/2023-03-30-vicuna/} {Vicuna: An
  open-source chatbot impressing gpt-4 with 90\%* chatgpt quality}.

\bibitem[{Chung et~al.(2022)Chung, Hou, Longpre, Zoph, Tay, Fedus, Li, Wang,
  Dehghani, Brahma et~al.}]{chung2022scaling}
Hyung~Won Chung, Le~Hou, Shayne Longpre, Barret Zoph, Yi~Tay, William Fedus,
  Eric Li, Xuezhi Wang, Mostafa Dehghani, Siddhartha Brahma, et~al. 2022.
\newblock Scaling instruction-finetuned language models.
\newblock \emph{arXiv preprint arXiv:2210.11416}.

\bibitem[{De-Arteaga et~al.(2019)De-Arteaga, Romanov, Wallach, Chayes, Borgs,
  Chouldechova, Geyik, Kenthapadi, and Kalai}]{10.1145/3287560.3287572}
Maria De-Arteaga, Alexey Romanov, Hanna Wallach, Jennifer Chayes, Christian
  Borgs, Alexandra Chouldechova, Sahin Geyik, Krishnaram Kenthapadi, and
  Adam~Tauman Kalai. 2019.
\newblock \href {https://doi.org/10.1145/3287560.3287572} {Bias in bios: A case
  study of semantic representation bias in a high-stakes setting}.
\newblock In \emph{Proceedings of the Conference on Fairness, Accountability,
  and Transparency}, FAT* '19, page 120–128, New York, NY, USA. Association
  for Computing Machinery.

\bibitem[{de~Vassimon~Manela et~al.(2021)de~Vassimon~Manela, Errington, Fisher,
  van Breugel, and Minervini}]{de-vassimon-manela-etal-2021-stereotype}
Daniel de~Vassimon~Manela, David Errington, Thomas Fisher, Boris van Breugel,
  and Pasquale Minervini. 2021.
\newblock \href {https://doi.org/10.18653/v1/2021.eacl-main.190} {Stereotype
  and skew: Quantifying gender bias in pre-trained and fine-tuned language
  models}.
\newblock In \emph{Proceedings of the 16th Conference of the European Chapter
  of the Association for Computational Linguistics: Main Volume}, pages
  2232--2242, Online. Association for Computational Linguistics.

\bibitem[{Delobelle et~al.(2021)Delobelle, Tokpo, Calders, and
  Berendt}]{delobelle2021measuring}
Pieter Delobelle, Ewoenam~Kwaku Tokpo, Toon Calders, and Bettina Berendt. 2021.
\newblock Measuring fairness with biased rulers: A survey on quantifying biases
  in pretrained language models.
\newblock \emph{arXiv preprint arXiv:2112.07447}.

\bibitem[{Devlin et~al.(2019{\natexlab{a}})Devlin, Chang, Lee, and
  Toutanova}]{devlin-etal-2019-bert}
Jacob Devlin, Ming-Wei Chang, Kenton Lee, and Kristina Toutanova.
  2019{\natexlab{a}}.
\newblock \href {https://doi.org/10.18653/v1/N19-1423} {{BERT}: Pre-training of
  deep bidirectional transformers for language understanding}.
\newblock In \emph{Proceedings of the 2019 Conference of the North {A}merican
  Chapter of the Association for Computational Linguistics: Human Language
  Technologies, Volume 1 (Long and Short Papers)}, pages 4171--4186,
  Minneapolis, Minnesota. Association for Computational Linguistics.

\bibitem[{Devlin et~al.(2019{\natexlab{b}})Devlin, Chang, Lee, and
  Toutanova}]{Devlin2019BERTPO}
Jacob Devlin, Ming-Wei Chang, Kenton Lee, and Kristina Toutanova.
  2019{\natexlab{b}}.
\newblock Bert: Pre-training of deep bidirectional transformers for language
  understanding.
\newblock \emph{ArXiv}, abs/1810.04805.

\bibitem[{Dodge et~al.(2021)Dodge, Sap, Marasovi{\'c}, Agnew, Ilharco,
  Groeneveld, Mitchell, and Gardner}]{dodge-etal-2021-documenting}
Jesse Dodge, Maarten Sap, Ana Marasovi{\'c}, William Agnew, Gabriel Ilharco,
  Dirk Groeneveld, Margaret Mitchell, and Matt Gardner. 2021.
\newblock \href {https://doi.org/10.18653/v1/2021.emnlp-main.98} {Documenting
  large webtext corpora: A case study on the colossal clean crawled corpus}.
\newblock In \emph{Proceedings of the 2021 Conference on Empirical Methods in
  Natural Language Processing}, pages 1286--1305, Online and Punta Cana,
  Dominican Republic. Association for Computational Linguistics.

\bibitem[{Dong et~al.(2022)Dong, Li, Dai, Zheng, Wu, Chang, Sun, Xu, and
  Sui}]{dong2022survey}
Qingxiu Dong, Lei Li, Damai Dai, Ce~Zheng, Zhiyong Wu, Baobao Chang, Xu~Sun,
  Jingjing Xu, and Zhifang Sui. 2022.
\newblock A survey for in-context learning.
\newblock \emph{arXiv preprint arXiv:2301.00234}.

\bibitem[{Foulds et~al.(2020)Foulds, Islam, Keya, and
  Pan}]{foulds2020intersectional}
James~R Foulds, Rashidul Islam, Kamrun~Naher Keya, and Shimei Pan. 2020.
\newblock An intersectional definition of fairness.
\newblock In \emph{2020 IEEE 36th International Conference on Data Engineering
  (ICDE)}, pages 1918--1921. IEEE.

\bibitem[{Gao et~al.(2021{\natexlab{a}})Gao, Fisch, and
  Chen}]{gao-etal-2021-making}
Tianyu Gao, Adam Fisch, and Danqi Chen. 2021{\natexlab{a}}.
\newblock \href {https://doi.org/10.18653/v1/2021.acl-long.295} {Making
  pre-trained language models better few-shot learners}.
\newblock In \emph{Proceedings of the 59th Annual Meeting of the Association
  for Computational Linguistics and the 11th International Joint Conference on
  Natural Language Processing (Volume 1: Long Papers)}, pages 3816--3830,
  Online. Association for Computational Linguistics.

\bibitem[{Gao et~al.(2021{\natexlab{b}})Gao, Fisch, and Chen}]{Gao2021MakingPL}
Tianyu Gao, Adam Fisch, and Danqi Chen. 2021{\natexlab{b}}.
\newblock Making pre-trained language models better few-shot learners.
\newblock \emph{ArXiv}, abs/2012.15723.

\bibitem[{Ghosh et~al.(2021)Ghosh, Genuit, and Reagan}]{pmlr-v142-ghosh21a}
Avijit Ghosh, Lea Genuit, and Mary Reagan. 2021.
\newblock \href {https://proceedings.mlr.press/v142/ghosh21a.html}
  {Characterizing intersectional group fairness with worst-case comparisons}.
\newblock In \emph{Proceedings of 2nd Workshop on Diversity in Artificial
  Intelligence (AIDBEI)}, volume 142 of \emph{Proceedings of Machine Learning
  Research}, pages 22--34. PMLR.

\bibitem[{Groenwold et~al.(2020)Groenwold, Ou, Parekh, Honnavalli, Levy, Mirza,
  and Wang}]{groenwold-etal-2020-investigating}
Sophie Groenwold, Lily Ou, Aesha Parekh, Samhita Honnavalli, Sharon Levy, Diba
  Mirza, and William~Yang Wang. 2020.
\newblock \href {https://doi.org/10.18653/v1/2020.emnlp-main.473}
  {Investigating {A}frican-{A}merican {V}ernacular {E}nglish in
  transformer-based text generation}.
\newblock In \emph{Proceedings of the 2020 Conference on Empirical Methods in
  Natural Language Processing (EMNLP)}, pages 5877--5883, Online. Association
  for Computational Linguistics.

\bibitem[{Han et~al.(2022)Han, Shen, Li, Frermann, Baldwin, and
  Cohn}]{han-etal-2022-fairlib}
Xudong Han, Aili Shen, Yitong Li, Lea Frermann, Timothy Baldwin, and Trevor
  Cohn. 2022.
\newblock \href {https://aclanthology.org/2022.emnlp-demos.7} {{F}air{L}ib: A
  unified framework for assessing and improving fairness}.
\newblock In \emph{Proceedings of the 2022 Conference on Empirical Methods in
  Natural Language Processing: System Demonstrations}, pages 60--71, Abu Dhabi,
  UAE. Association for Computational Linguistics.

\bibitem[{Hardt et~al.(2016)Hardt, Price, and Srebro}]{hardt2016equality}
Moritz Hardt, Eric Price, and Nati Srebro. 2016.
\newblock Equality of opportunity in supervised learning.
\newblock \emph{Advances in neural information processing systems}, 29.

\bibitem[{Holtzman et~al.(2021)Holtzman, West, Schwartz, Choi, and
  Zettlemoyer}]{Holtzman2021SurfaceFC}
Ari Holtzman, Peter West, Vered Schwartz, Yejin Choi, and Luke Zettlemoyer.
  2021.
\newblock Surface form competition: Why the highest probability answer isn’t
  always right.
\newblock \emph{ArXiv}, abs/2104.08315.

\bibitem[{Hovy(2015)}]{hovy-2015-demographic}
Dirk Hovy. 2015.
\newblock \href {https://doi.org/10.3115/v1/P15-1073} {Demographic factors
  improve classification performance}.
\newblock In \emph{Proceedings of the 53rd Annual Meeting of the Association
  for Computational Linguistics and the 7th International Joint Conference on
  Natural Language Processing (Volume 1: Long Papers)}, pages 752--762,
  Beijing, China. Association for Computational Linguistics.

\bibitem[{Islam et~al.(2021)Islam, Pan, and Foulds}]{islam2021can}
Rashidul Islam, Shimei Pan, and James~R Foulds. 2021.
\newblock Can we obtain fairness for free?
\newblock In \emph{Proceedings of the 2021 AAAI/ACM Conference on AI, Ethics,
  and Society}, pages 586--596.

\bibitem[{Jiang et~al.(2019)Jiang, Xu, Araki, and Neubig}]{Jiang2019HowCW}
Zhengbao Jiang, Frank~F. Xu, J.~Araki, and Graham Neubig. 2019.
\newblock How can we know what language models know?
\newblock \emph{Transactions of the Association for Computational Linguistics},
  8:423--438.

\bibitem[{Johnson et~al.(2019)Johnson, Douze, and
  J{\'e}gou}]{johnson2019billion}
Jeff Johnson, Matthijs Douze, and Herv{\'e} J{\'e}gou. 2019.
\newblock Billion-scale similarity search with {GPUs}.
\newblock \emph{IEEE Transactions on Big Data}, 7(3):535--547.

\bibitem[{Kaneko et~al.(2022)Kaneko, Bollegala, and
  Okazaki}]{kaneko-etal-2022-debiasing}
Masahiro Kaneko, Danushka Bollegala, and Naoaki Okazaki. 2022.
\newblock \href {https://aclanthology.org/2022.coling-1.111} {Debiasing isn{'}t
  enough! {--} on the effectiveness of debiasing {MLM}s and their social biases
  in downstream tasks}.
\newblock In \emph{Proceedings of the 29th International Conference on
  Computational Linguistics}, pages 1299--1310, Gyeongju, Republic of Korea.
  International Committee on Computational Linguistics.

\bibitem[{Kim(2022)}]{kim2022race}
Pauline~T Kim. 2022.
\newblock Race-aware algorithms: Fairness, nondiscrimination and affirmative
  action.
\newblock \emph{Cal. L. Rev.}, 110:1539.

\bibitem[{Kingma and Ba(2014)}]{kingma2014adam}
Diederik~P Kingma and Jimmy Ba. 2014.
\newblock Adam: A method for stochastic optimization.
\newblock \emph{arXiv preprint arXiv:1412.6980}.

\bibitem[{Kocielnik et~al.(2023)Kocielnik, Kangaslahti, Prabhumoye, Hari,
  Alvarez, and Anandkumar}]{pmlr-v203-kocielnik23a}
Rafal Kocielnik, Sara Kangaslahti, Shrimai Prabhumoye, Meena Hari, Michael
  Alvarez, and Anima Anandkumar. 2023.
\newblock \href {https://proceedings.mlr.press/v203/kocielnik23a.html} {Can you
  label less by using out-of-domain data? active and transfer learning with
  few-shot instructions}.
\newblock In \emph{Proceedings of The 1st Transfer Learning for Natural
  Language Processing Workshop}, volume 203 of \emph{Proceedings of Machine
  Learning Research}, pages 22--32. PMLR.

\bibitem[{Lewis et~al.(2019)Lewis, Liu, Goyal, Ghazvininejad, Mohamed, Levy,
  Stoyanov, and Zettlemoyer}]{Lewis2019BARTDS}
Mike Lewis, Yinhan Liu, Naman Goyal, Marjan Ghazvininejad, Abdelrahman Mohamed,
  Omer Levy, Veselin Stoyanov, and Luke Zettlemoyer. 2019.
\newblock Bart: Denoising sequence-to-sequence pre-training for natural
  language generation, translation, and comprehension.
\newblock In \emph{Annual Meeting of the Association for Computational
  Linguistics}.

\bibitem[{Lin et~al.(2022)Lin, Mihaylov, Artetxe, Wang, Chen, Simig, Ott,
  Goyal, Bhosale, Du, Pasunuru, Shleifer, Koura, Chaudhary, O{'}Horo, Wang,
  Zettlemoyer, Kozareva, Diab, Stoyanov, and Li}]{lin-etal-2022-shot}
Xi~Victoria Lin, Todor Mihaylov, Mikel Artetxe, Tianlu Wang, Shuohui Chen,
  Daniel Simig, Myle Ott, Naman Goyal, Shruti Bhosale, Jingfei Du, Ramakanth
  Pasunuru, Sam Shleifer, Punit~Singh Koura, Vishrav Chaudhary, Brian O{'}Horo,
  Jeff Wang, Luke Zettlemoyer, Zornitsa Kozareva, Mona Diab, Veselin Stoyanov,
  and Xian Li. 2022.
\newblock \href {https://aclanthology.org/2022.emnlp-main.616} {Few-shot
  learning with multilingual generative language models}.
\newblock In \emph{Proceedings of the 2022 Conference on Empirical Methods in
  Natural Language Processing}, pages 9019--9052, Abu Dhabi, United Arab
  Emirates. Association for Computational Linguistics.

\bibitem[{Liu et~al.(2021)Liu, Shen, Zhang, Dolan, Carin, and
  Chen}]{Liu2021WhatMG}
Jiachang Liu, Dinghan Shen, Yizhe Zhang, Bill Dolan, Lawrence Carin, and Weizhu
  Chen. 2021.
\newblock What makes good in-context examples for gpt-3?
\newblock In \emph{Workshop on Knowledge Extraction and Integration for Deep
  Learning Architectures; Deep Learning Inside Out}.

\bibitem[{Liu et~al.(2022)Liu, Shen, Zhang, Dolan, Carin, and
  Chen}]{liu-etal-2022-makes}
Jiachang Liu, Dinghan Shen, Yizhe Zhang, Bill Dolan, Lawrence Carin, and Weizhu
  Chen. 2022.
\newblock \href {https://doi.org/10.18653/v1/2022.deelio-1.10} {What makes good
  in-context examples for {GPT}-3?}
\newblock In \emph{Proceedings of Deep Learning Inside Out (DeeLIO 2022): The
  3rd Workshop on Knowledge Extraction and Integration for Deep Learning
  Architectures}, pages 100--114, Dublin, Ireland and Online. Association for
  Computational Linguistics.

\bibitem[{Liu et~al.(2019{\natexlab{a}})Liu, Ott, Goyal, Du, Joshi, Chen, Levy,
  Lewis, Zettlemoyer, and Stoyanov}]{DBLP:journals/corr/abs-1907-11692}
Yinhan Liu, Myle Ott, Naman Goyal, Jingfei Du, Mandar Joshi, Danqi Chen, Omer
  Levy, Mike Lewis, Luke Zettlemoyer, and Veselin Stoyanov. 2019{\natexlab{a}}.
\newblock \href {http://arxiv.org/abs/1907.11692} {Roberta: {A} robustly
  optimized {BERT} pretraining approach}.
\newblock \emph{CoRR}, abs/1907.11692.

\bibitem[{Liu et~al.(2019{\natexlab{b}})Liu, Ott, Goyal, Du, Joshi, Chen, Levy,
  Lewis, Zettlemoyer, and Stoyanov}]{Liu2019RoBERTaAR}
Yinhan Liu, Myle Ott, Naman Goyal, Jingfei Du, Mandar Joshi, Danqi Chen, Omer
  Levy, Mike Lewis, Luke Zettlemoyer, and Veselin Stoyanov. 2019{\natexlab{b}}.
\newblock Roberta: A robustly optimized bert pretraining approach.
\newblock \emph{ArXiv}, abs/1907.11692.

\bibitem[{Lu et~al.(2021)Lu, Bartolo, Moore, Riedel, and
  Stenetorp}]{Lu2021FantasticallyOP}
Yao Lu, Max Bartolo, Alastair Moore, Sebastian Riedel, and Pontus Stenetorp.
  2021.
\newblock Fantastically ordered prompts and where to find them: Overcoming
  few-shot prompt order sensitivity.
\newblock In \emph{Annual Meeting of the Association for Computational
  Linguistics}.

\bibitem[{Lu et~al.(2022)Lu, Bartolo, Moore, Riedel, and
  Stenetorp}]{lu-etal-2022-fantastically}
Yao Lu, Max Bartolo, Alastair Moore, Sebastian Riedel, and Pontus Stenetorp.
  2022.
\newblock \href {https://doi.org/10.18653/v1/2022.acl-long.556} {Fantastically
  ordered prompts and where to find them: Overcoming few-shot prompt order
  sensitivity}.
\newblock In \emph{Proceedings of the 60th Annual Meeting of the Association
  for Computational Linguistics (Volume 1: Long Papers)}, pages 8086--8098,
  Dublin, Ireland. Association for Computational Linguistics.

\bibitem[{Ma et~al.(2023)Ma, Zhang, Bian, Liu, Zhang, Zhao, Zhang, Fu, Hu, and
  Wu}]{ma2023fairness}
Huan Ma, Changqing Zhang, Yatao Bian, Lemao Liu, Zhirui Zhang, Peilin Zhao, Shu
  Zhang, Huazhu Fu, Qinghua Hu, and Bingzhe Wu. 2023.
\newblock Fairness-guided few-shot prompting for large language models.
\newblock \emph{arXiv preprint arXiv:2303.13217}.

\bibitem[{Magee et~al.(2021)Magee, Ghahremanlou, Soldatic, and
  Robertson}]{magee2021intersectional}
Liam Magee, Lida Ghahremanlou, Karen Soldatic, and Shanthi Robertson. 2021.
\newblock Intersectional bias in causal language models.
\newblock \emph{arXiv preprint arXiv:2107.07691}.

\bibitem[{Mathew et~al.(2021)Mathew, Saha, Yimam, Biemann, Goyal, and
  Mukherjee}]{mathew2021hatexplain}
Binny Mathew, Punyajoy Saha, Seid~Muhie Yimam, Chris Biemann, Pawan Goyal, and
  Animesh Mukherjee. 2021.
\newblock Hatexplain: A benchmark dataset for explainable hate speech
  detection.
\newblock In \emph{Proceedings of the AAAI Conference on Artificial
  Intelligence}, volume~35, pages 14867--14875.

\bibitem[{Min et~al.(2021{\natexlab{a}})Min, Lewis, Hajishirzi, and
  Zettlemoyer}]{Min2021NoisyCL}
Sewon Min, Michael Lewis, Hannaneh Hajishirzi, and Luke Zettlemoyer.
  2021{\natexlab{a}}.
\newblock Noisy channel language model prompting for few-shot text
  classification.
\newblock In \emph{Annual Meeting of the Association for Computational
  Linguistics}.

\bibitem[{Min et~al.(2021{\natexlab{b}})Min, Lewis, Zettlemoyer, and
  Hajishirzi}]{Min2021MetaICLLT}
Sewon Min, Mike Lewis, Luke Zettlemoyer, and Hannaneh Hajishirzi.
  2021{\natexlab{b}}.
\newblock Metaicl: Learning to learn in context.
\newblock \emph{ArXiv}, abs/2110.15943.

\bibitem[{Min et~al.(2022)Min, Lyu, Holtzman, Artetxe, Lewis, Hajishirzi, and
  Zettlemoyer}]{min-etal-2022-rethinking}
Sewon Min, Xinxi Lyu, Ari Holtzman, Mikel Artetxe, Mike Lewis, Hannaneh
  Hajishirzi, and Luke Zettlemoyer. 2022.
\newblock \href {https://aclanthology.org/2022.emnlp-main.759} {Rethinking the
  role of demonstrations: What makes in-context learning work?}
\newblock In \emph{Proceedings of the 2022 Conference on Empirical Methods in
  Natural Language Processing}, pages 11048--11064, Abu Dhabi, United Arab
  Emirates. Association for Computational Linguistics.

\bibitem[{Nadeem et~al.(2021)Nadeem, Bethke, and
  Reddy}]{nadeem-etal-2021-stereoset}
Moin Nadeem, Anna Bethke, and Siva Reddy. 2021.
\newblock \href {https://doi.org/10.18653/v1/2021.acl-long.416} {{S}tereo{S}et:
  Measuring stereotypical bias in pretrained language models}.
\newblock In \emph{Proceedings of the 59th Annual Meeting of the Association
  for Computational Linguistics and the 11th International Joint Conference on
  Natural Language Processing (Volume 1: Long Papers)}, pages 5356--5371,
  Online. Association for Computational Linguistics.

\bibitem[{Nangia et~al.(2020)Nangia, Vania, Bhalerao, and
  Bowman}]{nangia-etal-2020-crows}
Nikita Nangia, Clara Vania, Rasika Bhalerao, and Samuel~R. Bowman. 2020.
\newblock \href {https://doi.org/10.18653/v1/2020.emnlp-main.154}
  {{C}row{S}-pairs: A challenge dataset for measuring social biases in masked
  language models}.
\newblock In \emph{Proceedings of the 2020 Conference on Empirical Methods in
  Natural Language Processing (EMNLP)}, pages 1953--1967, Online. Association
  for Computational Linguistics.

\bibitem[{Nguyen et~al.(2020)Nguyen, Vu, and
  Tuan~Nguyen}]{nguyen-etal-2020-bertweet}
Dat~Quoc Nguyen, Thanh Vu, and Anh Tuan~Nguyen. 2020.
\newblock \href {https://doi.org/10.18653/v1/2020.emnlp-demos.2} {{BERT}weet: A
  pre-trained language model for {E}nglish tweets}.
\newblock In \emph{Proceedings of the 2020 Conference on Empirical Methods in
  Natural Language Processing: System Demonstrations}, pages 9--14, Online.
  Association for Computational Linguistics.

\bibitem[{Ouyang et~al.(2022)Ouyang, Wu, Jiang, Almeida, Wainwright, Mishkin,
  Zhang, Agarwal, Slama, Ray et~al.}]{ouyang2022training}
Long Ouyang, Jeffrey Wu, Xu~Jiang, Diogo Almeida, Carroll Wainwright, Pamela
  Mishkin, Chong Zhang, Sandhini Agarwal, Katarina Slama, Alex Ray, et~al.
  2022.
\newblock Training language models to follow instructions with human feedback.
\newblock \emph{Advances in Neural Information Processing Systems},
  35:27730--27744.

\bibitem[{Perez et~al.(2021)Perez, Kiela, and Cho}]{perez2021true}
Ethan Perez, Douwe Kiela, and Kyunghyun Cho. 2021.
\newblock True few-shot learning with language models.
\newblock \emph{Advances in neural information processing systems},
  34:11054--11070.

\bibitem[{Phang et~al.(2020)Phang, Yeres, Swanson, Liu, Tenney, Htut, Vania,
  Wang, and Bowman}]{phang2020jiant}
Jason Phang, Phil Yeres, Jesse Swanson, Haokun Liu, Ian~F. Tenney, Phu~Mon
  Htut, Clara Vania, Alex Wang, and Samuel~R. Bowman. 2020.
\newblock \texttt{jiant} 2.0: A software toolkit for research on
  general-purpose text understanding models.
\newblock \url{http://jiant.info/}.

\bibitem[{Portillo~Wightman et~al.(2023)Portillo~Wightman, DeLucia, and
  Dredze}]{portillo-etal-2023-strength}
Gwenyth Portillo~Wightman, Alexandra DeLucia, and Mark Dredze. 2023.
\newblock Strength in numbers: Estimating confidence of large language models
  by prompt agreement.
\newblock In \emph{Proceedings of the 3rd Workshop on Trustworthy Natural
  Language Processing (TrustNLP 2023)}, Toronto, CA. Association for
  Computational Linguistics.

\bibitem[{Radford et~al.(2019)Radford, Wu, Child, Luan, Amodei, and
  Sutskever}]{Radford2019LanguageMA}
Alec Radford, Jeff Wu, Rewon Child, David Luan, Dario Amodei, and Ilya
  Sutskever. 2019.
\newblock Language models are unsupervised multitask learners.

\bibitem[{Reimers and Gurevych(2019)}]{reimers-gurevych-2019-sentence}
Nils Reimers and Iryna Gurevych. 2019.
\newblock \href {https://doi.org/10.18653/v1/D19-1410} {Sentence-{BERT}:
  Sentence embeddings using {S}iamese {BERT}-networks}.
\newblock In \emph{Proceedings of the 2019 Conference on Empirical Methods in
  Natural Language Processing and the 9th International Joint Conference on
  Natural Language Processing (EMNLP-IJCNLP)}, pages 3982--3992, Hong Kong,
  China. Association for Computational Linguistics.

\bibitem[{Rubin et~al.(2021)Rubin, Herzig, and Berant}]{Rubin2021LearningTR}
Ohad Rubin, Jonathan Herzig, and Jonathan Berant. 2021.
\newblock Learning to retrieve prompts for in-context learning.
\newblock \emph{ArXiv}, abs/2112.08633.

\bibitem[{Salewski et~al.(2023)Salewski, Alaniz, Rio-Torto, Schulz, and
  Akata}]{salewski2023context}
Leonard Salewski, Stephan Alaniz, Isabel Rio-Torto, Eric Schulz, and Zeynep
  Akata. 2023.
\newblock In-context impersonation reveals large language models' strengths and
  biases.
\newblock \emph{arXiv preprint arXiv:2305.14930}.

\bibitem[{Shen et~al.(2022)Shen, Han, Cohn, Baldwin, and
  Frermann}]{shen-etal-2022-optimising}
Aili Shen, Xudong Han, Trevor Cohn, Timothy Baldwin, and Lea Frermann. 2022.
\newblock \href {https://doi.org/10.18653/v1/2022.naacl-main.299} {Optimising
  equal opportunity fairness in model training}.
\newblock In \emph{Proceedings of the 2022 Conference of the North American
  Chapter of the Association for Computational Linguistics: Human Language
  Technologies}, pages 4073--4084, Seattle, United States. Association for
  Computational Linguistics.

\bibitem[{Sheng et~al.(2019)Sheng, Chang, Natarajan, and
  Peng}]{sheng-etal-2019-woman}
Emily Sheng, Kai-Wei Chang, Premkumar Natarajan, and Nanyun Peng. 2019.
\newblock \href {https://doi.org/10.18653/v1/D19-1339} {The woman worked as a
  babysitter: On biases in language generation}.
\newblock In \emph{Proceedings of the 2019 Conference on Empirical Methods in
  Natural Language Processing and the 9th International Joint Conference on
  Natural Language Processing (EMNLP-IJCNLP)}, pages 3407--3412, Hong Kong,
  China. Association for Computational Linguistics.

\bibitem[{Stanczak and Augenstein(2021)}]{stanczak2021survey}
Karolina Stanczak and Isabelle Augenstein. 2021.
\newblock A survey on gender bias in natural language processing.
\newblock \emph{arXiv preprint arXiv:2112.14168}.

\bibitem[{Tal et~al.(2022)Tal, Magar, and Schwartz}]{tal-etal-2022-fewer}
Yarden Tal, Inbal Magar, and Roy Schwartz. 2022.
\newblock \href {https://doi.org/10.18653/v1/2022.gebnlp-1.13} {Fewer errors,
  but more stereotypes? the effect of model size on gender bias}.
\newblock In \emph{Proceedings of the 4th Workshop on Gender Bias in Natural
  Language Processing (GeBNLP)}, pages 112--120, Seattle, Washington.
  Association for Computational Linguistics.

\bibitem[{Taori et~al.(2023)Taori, Gulrajani, Zhang, Dubois, Li, Guestrin,
  Liang, and Hashimoto}]{alpaca}
Rohan Taori, Ishaan Gulrajani, Tianyi Zhang, Yann Dubois, Xuechen Li, Carlos
  Guestrin, Percy Liang, and Tatsunori~B. Hashimoto. 2023.
\newblock Stanford alpaca: An instruction-following llama model.
\newblock \url{https://github.com/tatsu-lab/stanford_alpaca}.

\bibitem[{Tay et~al.(2023)Tay, Dehghani, Tran, Garcia, Wei, Wang, Chung, Bahri,
  Schuster, Zheng, Zhou, Houlsby, and Metzler}]{tay2023ul}
Yi~Tay, Mostafa Dehghani, Vinh~Q. Tran, Xavier Garcia, Jason Wei, Xuezhi Wang,
  Hyung~Won Chung, Dara Bahri, Tal Schuster, Steven Zheng, Denny Zhou, Neil
  Houlsby, and Donald Metzler. 2023.
\newblock \href {https://openreview.net/forum?id=6ruVLB727MC} {{UL}2: Unifying
  language learning paradigms}.
\newblock In \emph{The Eleventh International Conference on Learning
  Representations}.

\bibitem[{Touvron et~al.(2023{\natexlab{a}})Touvron, Lavril, Izacard, Martinet,
  Lachaux, Lacroix, Rozi{\`e}re, Goyal, Hambro, Azhar
  et~al.}]{touvron2023llama}
Hugo Touvron, Thibaut Lavril, Gautier Izacard, Xavier Martinet, Marie-Anne
  Lachaux, Timoth{\'e}e Lacroix, Baptiste Rozi{\`e}re, Naman Goyal, Eric
  Hambro, Faisal Azhar, et~al. 2023{\natexlab{a}}.
\newblock Llama: Open and efficient foundation language models.
\newblock \emph{arXiv preprint arXiv:2302.13971}.

\bibitem[{Touvron et~al.(2023{\natexlab{b}})Touvron, Martin, Stone, Albert,
  Almahairi, Babaei, Bashlykov, Batra, Bhargava, Bhosale
  et~al.}]{touvron2023llama2}
Hugo Touvron, Louis Martin, Kevin Stone, Peter Albert, Amjad Almahairi, Yasmine
  Babaei, Nikolay Bashlykov, Soumya Batra, Prajjwal Bhargava, Shruti Bhosale,
  et~al. 2023{\natexlab{b}}.
\newblock Llama 2: Open foundation and fine-tuned chat models.
\newblock \emph{arXiv preprint arXiv:2307.09288}.

\bibitem[{Wang et~al.(2022)Wang, Kordi, Mishra, Liu, Smith, Khashabi, and
  Hajishirzi}]{wang2022self}
Yizhong Wang, Yeganeh Kordi, Swaroop Mishra, Alisa Liu, Noah~A Smith, Daniel
  Khashabi, and Hannaneh Hajishirzi. 2022.
\newblock Self-instruct: Aligning language model with self generated
  instructions.
\newblock \emph{arXiv preprint arXiv:2212.10560}.

\bibitem[{Wei et~al.(2022)Wei, Wang, Schuurmans, Bosma, Chi, Le, and
  Zhou}]{Wei2022ChainOT}
Jason Wei, Xuezhi Wang, Dale Schuurmans, Maarten Bosma, Ed~Chi, Quoc Le, and
  Denny Zhou. 2022.
\newblock Chain of thought prompting elicits reasoning in large language
  models.
\newblock \emph{ArXiv}, abs/2201.11903.

\bibitem[{Ye et~al.(2021)Ye, Lin, and Ren}]{ye-etal-2021-crossfit}
Qinyuan Ye, Bill~Yuchen Lin, and Xiang Ren. 2021.
\newblock \href {https://doi.org/10.18653/v1/2021.emnlp-main.572}
  {{C}ross{F}it: A few-shot learning challenge for cross-task generalization in
  {NLP}}.
\newblock In \emph{Proceedings of the 2021 Conference on Empirical Methods in
  Natural Language Processing}, pages 7163--7189, Online and Punta Cana,
  Dominican Republic. Association for Computational Linguistics.

\bibitem[{Ye et~al.(2023)Ye, Ou, Li, Ma, Yanggong, Wu, Fu, Chen, Zhao
  et~al.}]{ye2023assessing}
Wentao Ye, Mingfeng Ou, Tianyi Li, Xuetao Ma, Yifan Yanggong, Sai Wu, Jie Fu,
  Gang Chen, Junbo Zhao, et~al. 2023.
\newblock Assessing hidden risks of llms: An empirical study on robustness,
  consistency, and credibility.
\newblock \emph{arXiv preprint arXiv:2305.10235}.

\bibitem[{Zhang et~al.(2022{\natexlab{a}})Zhang, Zhang, Zhang, Fan, Li, Liu,
  and Chang}]{zhang2022fairness}
Guanhua Zhang, Yihua Zhang, Yang Zhang, Wenqi Fan, Qing Li, Sijia Liu, and
  Shiyu Chang. 2022{\natexlab{a}}.
\newblock Fairness reprogramming.
\newblock \emph{arXiv preprint arXiv:2209.10222}.

\bibitem[{Zhang et~al.(2020)Zhang, Lu, Abdalla, McDermott, and
  Ghassemi}]{zhang2020hurtful}
Haoran Zhang, Amy~X Lu, Mohamed Abdalla, Matthew McDermott, and Marzyeh
  Ghassemi. 2020.
\newblock Hurtful words: quantifying biases in clinical contextual word
  embeddings.
\newblock In \emph{proceedings of the ACM Conference on Health, Inference, and
  Learning}, pages 110--120.

\bibitem[{Zhang et~al.(2022{\natexlab{b}})Zhang, Zhang, Li, and
  Smola}]{zhang2022automatic}
Zhuosheng Zhang, Aston Zhang, Mu~Li, and Alex Smola. 2022{\natexlab{b}}.
\newblock Automatic chain of thought prompting in large language models.
\newblock \emph{arXiv preprint arXiv:2210.03493}.

\bibitem[{Zhao et~al.(2021)Zhao, Wallace, Feng, Klein, and
  Singh}]{Zhao2021CalibrateBU}
Tony Zhao, Eric Wallace, Shi Feng, Dan Klein, and Sameer Singh. 2021.
\newblock Calibrate before use: Improving few-shot performance of language
  models.
\newblock \emph{ArXiv}, abs/2102.09690.

\end{thebibliography}
